\documentclass[runningheads]{llncs}

\usepackage{eccv}

\usepackage{eccvabbrv}
\usepackage{hyperref}
\usepackage{graphicx}
\usepackage{booktabs}
\usepackage{float}
\usepackage{amsmath}
\usepackage{amssymb}
\usepackage{booktabs}
\usepackage{mathrsfs}
\usepackage{tabularx}
\usepackage{multirow}
\usepackage{pifont}
\usepackage{arydshln}
\usepackage{makecell}
\usepackage[percent]{overpic}
\usepackage[capitalize]{cleveref}
\usepackage{pifont}
\crefname{section}{Sec.}{Secs.}
\Crefname{section}{Section}{Sections}
\usepackage{placeins} 
\Crefname{table}{Table}{Tables}
\crefname{table}{Tab.}{Tabs.}
\usepackage[accsupp]{axessibility}  

\usepackage{orcidlink}

\begin{document}

\title{PASTA: Towards Flexible and Efficient HDR Imaging Via Progressively Aggregated Spatio-Temporal Alignment
} 

\titlerunning{PASTA}

\author{Xiaoning Liu\inst{1} \and
Ao Li\inst{1} \and
Zongwei Wu\inst{2} \and
Yapeng Du \inst{1} \and
Le Zhang\inst{1} \and \\
Yulun Zhang \inst{3} \and
Radu Timofte\inst{2,4} \and
Ce Zhu \inst{1}}

\authorrunning{X.~Liu et al.}

\institute{University of Electronic Science and Technology of China \and
University of Würzburg \and
Shanghai Jiao Tong University \and
ETH Zurich}

\maketitle

\begin{abstract}
  Leveraging Transformer attention has led to great advancements in HDR deghosting. However, the intricate nature of self-attention introduces practical challenges, as existing state-of-the-art methods often demand high-end GPUs or exhibit slow inference speeds, especially for high-resolution images like 2K. Striking an optimal balance between performance and latency remains a critical concern. In response, this work presents PASTA, a novel Progressively Aggregated Spatio-Temporal Alignment framework for HDR deghosting. Our approach achieves effectiveness and efficiency by harnessing hierarchical representation during feature distanglement. Through the utilization of diverse granularities within the hierarchical structure, our method substantially boosts computational speed and optimizes the HDR imaging workflow. In addition, we explore within-scale feature modeling with local and global attention, gradually merging and refining them in a coarse-to-fine fashion. Experimental results showcase PASTA's superiority over current SOTA methods in both visual quality and performance metrics, accompanied by a substantial 3-fold ($\times 3$) increase in inference speed. 
\end{abstract}

\section{Introduction}
\label{sec:intro}
Imaging sensors' limited dynamic range often compromises photo quality by losing detail in highlights and shadows. Multi-frame HDR imaging, which involves capturing a stack of low dynamic range (LDR) images at different exposures and then combining them, addresses this by expanding the dynamic range and enhancing visual quality. However, this technique faces challenges in accurately aligning frames in the presence of movement and shake, as well as effectively merging frames of varying exposures while dealing with occlusion.

Recent advancements in HDR imaging have evolved from early methods using hand-crafted features \cite{jacobs2008automatic, lu2009high, kang2003high, myszkowski2022high, hu2013hdr, sen2012robust} to contemporary learning-based models. These models excel in cross-exposure mappings due to inductive bias \cite{kalantari2017deep, wu2018deep, yan2020deep, yan2019attention, marin2022drhdr, li2022gamma}, with Transformer models further enhancing HDR imaging by introducing contextualized awareness \cite{liu2022ghost, song2022selective, yan2023unified, liu2023joint, tel2023alignment}.

Despite these advancements, a major hurdle is the rising computational complexity with higher image resolutions. The self-attention mechanism in Transformers requires significant computational resources and memory, limiting their real-world applicability, especially with the common use of high-resolution images. While methods like \cite{tel2022cen, prabhakar2020towards} can process 2K images, this often comes at the expense of performance.
\begin{figure}[t]
\centering
    \begin{overpic}[width=0.8\linewidth,tics=5]{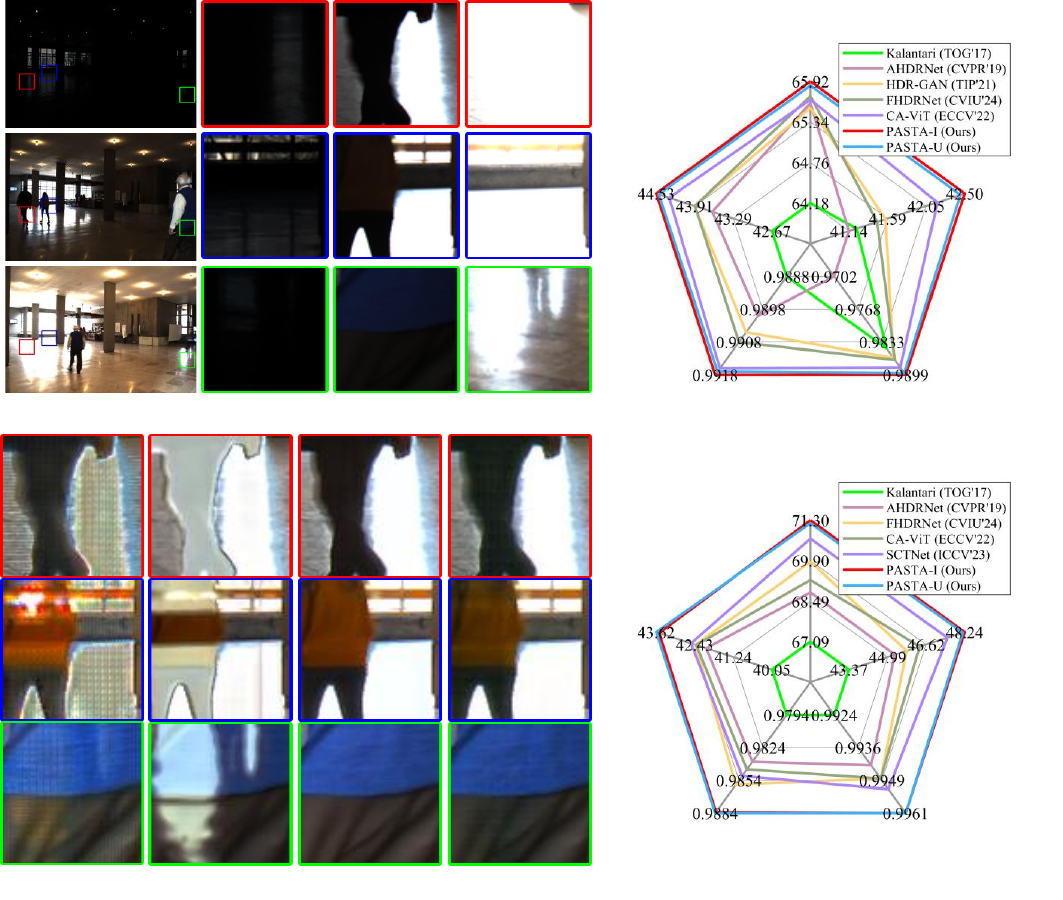}
        \put(10, -2){\scriptsize{(a) Tursun \etal's dataset \cite{tursun2015state}}}
    	\put(58, 43){\scriptsize{(b) Kalantari~\etal's~\cite{kalantari2017deep} dataset}}
    	\put(60, -2){\scriptsize{(c) Tel~\etal's~\cite{tel2023alignment} dataset}}
        \put(56, 68){\tiny{PSNR-$\mu$}}
        \put(90, 68){\tiny{PSNR-$\emph{l}$}}
        \put(73, 78){\tiny{H.V.-2}}
        \put(83, 46.5){\tiny{SSIM-$\emph{l}$}}
        \put(63, 46.5){\tiny{SSIM-$\mu$}}
        
        \put(56, 27){\tiny{PSNR-$\mu$}}
        \put(90, 27){\tiny{PSNR-$\emph{l}$}}
        \put(73, 37){\tiny{H.V.-2}}
        \put(83, 5){\tiny{SSIM-$\emph{l}$}}
        \put(63, 5){\tiny{SSIM-$\mu$}}

        \put(6, 45.5){\tiny{LDRs}}
        \put(0.3, 1){\tiny{NHDRRnet}}
        \put(17, 1){\tiny{CA-ViT}}
        \put(30, 1){\tiny{PASTA-I}}
        \put(44, 1){\tiny{PASTA-U}}
        \put(29.5, 45.5){\tiny{LDRs Patches}}

    \end{overpic}
\caption{\textbf{Visual and quantitative comparison.} Our method consistently outperforms in two benchmark HDR datasets, urpassing the state-of-the-art CA-ViT \cite{liu2022ghost} and NHDRRnet \cite{yan2020deep} (the hierarchical method) methods in preserving content integrity, achieving ghost-free and high-fidelity results. Zoom in to see more details.}
\label{fig:exampe_intro}
\vspace{-4mm}
\end{figure}
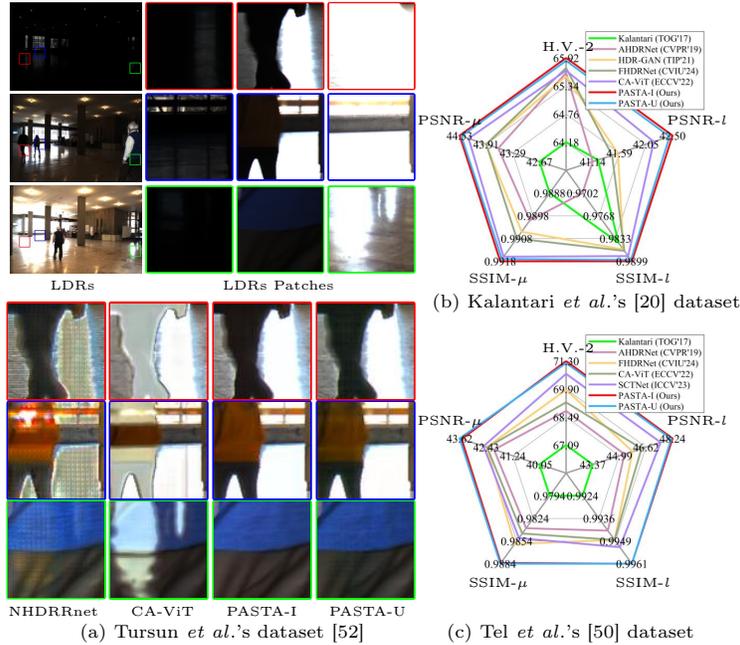

\begin{table}[!t]
\centering
\caption{\textbf{Components in existing HDR deghosting methods}. Components are categorized by functionality: i) \textit{Alignment} involves the spatial transformation of misaligned images or features, including homography transformation (HT), spatial attention (SA), pyramid cross attention (PCA), patch aggregation (PC) and temporal attention (TA); ii) \textit{Aggregation} characterizes the architecture of feature aggregation; iii) \textit{Domain} specifies aggregation in either the spatial or frequency domain, and iv)\textit{
High Resolution (HR)} refers to the capability to handle 2K images.}
\vspace{-2.5mm}
\resizebox{\textwidth}{!}{
    \begin{tabular}{c|c|c|c|c|c|c|c|c|c|c}
        \hline
        \multirow{2}{*}{} & 
        \multicolumn{4}{c|}{CNN-based Methods} & \multicolumn{6}{c}{Transformer-based Methods} \\ 
        \cline{2-11} 
        & \scriptsize{DeepHDR \cite{wu2018deep}} 
        & \scriptsize{AHDRNet \cite{yan2019attention}} 
        & \scriptsize{HDR-GAN \cite{niu2021hdr}}
        & \scriptsize{FHDRNet \cite{dai2024wavelet}}
        & \scriptsize{CA-ViT \cite{liu2022ghost}}
        & \scriptsize{Joint-HDRDN \cite{liu2023joint}} 
        & \scriptsize{HyHDRNet \cite{yan2023unified}} 
        & \scriptsize{SCTNet \cite{tel2023alignment}} 
        & \scriptsize{PASTA-I} 
        & \scriptsize{PASTA-U} \\ 
        & \scriptsize{(ECCV'18)}
        & \scriptsize{(CVPR'19)}
        & \scriptsize{(TIP'21)}
        & \scriptsize{(CVIU'24)}
        & \scriptsize{(ECCV'22)}
        & \scriptsize{(CVPR'23)}
        & \scriptsize{(CVPR'23)}
        & \scriptsize{(ICCV'23)}
        & \scriptsize{(Ours)}
        & \scriptsize{(Ours)} \\
        \hline
        Alignment & Yes (HT) & Yes (SA) & No & No & Yes (SA) & Yes (PCA) & Yes (PC \& SA) & No & Yes (TA) & No \\
        Aggregation & Hierarchical & Plain & Hierarchical & Hierarchical & Plain & Plain & Plain & Plain &
        Hierarchical & Hierarchical \\
        Domain & Spat. & Spat. & Spat. & Freq. & Spat. & Spat. & Spat. & Spat. & Spat. \& Freq. & Spat. \& Freq.  \\ 
        HR (2K) & Yes & No & Yes & Yes & No & No & No & No & Yes & Yes \\
        \hline   
    \end{tabular}}
\vspace{-2.0em}
\label{tab:components_HDR}
\end{table}
This necessitates further reflection and reevaluation of HDR deghosting methods, aiming to construct a simple, powerful and efficient framework. Our summaries, detailed in \cref{tab:components_HDR}, distill the methodology into four key components: \textit{Alignment}, \textit{Aggregation}, \textit{Domain}, and \textit{High Resolution}, each critical to the practical effectiveness of HDR deghosting approaches.

We discern that hierarchical representation presents a natural solution capable of simultaneously addressing multiple challenges. This representation inherently contains information from different levels, making it a robust candidate for our objectives. Moreover, hierarchical representation coupled with diverse granularities aligns well with computational efficiency, further enhancing its appeal for our purposes. Motivated by these considerations, we propose a novel framework named Progressively Aggregated Spatio-Temporal Alignment (PASTA) designed to fully exploit the benefits of hierarchical representation for HDR imaging. Inspired by the effectiveness of wavelet statistical modeling \cite{crouse1998wavelet}, we utilize the discrete wavelet transform (DWT) \cite{mallat1989theory} to establish a hierarchical representation framework. Although DWT is commonly employed in various low-level vision tasks \cite{ramamonjisoa2021single,yu2021wavefill,dai2024wavelet}, prior works predominantly focus on utilizing it to compute frequency-domain features and just perform image-frequency fusion, overlooking the statistical correlation of wavelet coefficients. In contrast, we innovatively incorporate channel attention and spatial self-attention to implicitly capture inter- and intra-scale relationships among subbands. Our approach explores the potential of DWT as a hierarchical representation, offering advantages such as inter-/cross-scale interactions, flexibility, improved computational efficiency and suitability, which is the first time addressed for high-resolution HDR deghosting.



Still, an additional challenge emerges: how to efficiently extract the most informative features and establish cross-hierarchy interactions for effective fusion of all available information. To meet this challenge, we devise a gradual fusion strategy. Initially, we undertake an in-depth analysis incorporating global and local attention mechanisms. Subsequently, we systematically progress to the fusion of information across different hierarchies. This progressive approach aims to capture nuanced details and global context, ensuring the comprehensive utilization of our hierarchical representation.

As shown in \cref{fig:exampe_intro}, PASTA sets new SOTA records on HDR benchmarks, validating its effectiveness and versatility. It also excels in efficiency, with fast inference speeds, low GPU memory usage, and reduced latency. Remarkably, PASTA outperforms existing SOTA methods in processing 1080P LDR images on a plain GPU with just 12GB of RAM, achieving a substantial 3-fold ($\times3$) increase in inference speed. An ultra-lightweight version further enhances this, boosting speed by nearly 9 times ($\times9$) while still maintaining a competitive level of performance.

\vspace{-4mm}
\section{Related Work}
\vspace{-3mm}
\noindent{\bf Rejection-based methods.} These methods aim to identify and reject moving regions inconsistent with a reference frame's background during synthesis. Various techniques, such as non-parametric estimation \cite{khan2006ghost}, local entropy \cite{jacobs2008automatic}, threshold bitmap \cite{pece2010bitmap, lu2009high}, intensity histogram \cite{min2009histogram}, and bi-directional similarity \cite{zheng2013hybrid, li2014selectively}, detect moving pixels. However, they often lose fine details in dynamic areas and struggle to adapt to different cameras and exposure settings due to fixed threshold reliance \cite{granados2013automatic}.

\noindent{\bf Registration-based methods.} The emphasis is very much on aligning exposure images. In the case of only camera motion between exposure sources, \ie, the captured scene is rigid and on a plane, simple operations such as translation \cite{ward2003fast}, homography \cite{tomaszewska2007image}, correlation matching \cite{akyuz2011photographically}, and phase cross correlation \cite{yao2011robust} can be used to crack this challenge. For dynamic scenes, most methods seek non-grid dense correspondences (\eg, joint homography and SIFT \cite{heo2010ghost}, joint homography and optical flow \cite{kang2003high}, and RANSAC \cite{hu2012exposure, gallo2015locally}) between images.
However, the warped images are prone to holes in the presence of large displacements or severe occlusion.

\noindent{\bf Optimization-based methods.} They jointly perform alignment and reconstruction in the form of an energy function, or only use the framework to perform one of them, for example, using Poisson equation \cite{fattal2002gradient, gallo2009artifact}, graph cut \cite{eden2006seamless}, Markov random field \cite{jinno2008motion, granados2013automatic}, Bayesian \cite{lu2009high}, energy-based optic flow \cite{zimmer2011freehand}, PatchMatch \cite{sen2012robust, hu2013hdr}, rank minimization \cite{lee2014ghost, oh2014robust} and background estimation \cite{granados2008background}. 
Despite progress, they fall short of learning-based methods in deghosting results and tend to be time-consuming.

\noindent{\bf Learning-based methods.} They have evolved since \cite{kalantari2017deep}, which uses optical flow \cite{liu2009beyond} for image alignment in HDR creation. Subsequent works directly cascaded features \cite{wu2018deep, niu2021hdr, ye2021progressive} or utilize attention mechanisms \cite{yan2019attention, marin2022drhdr, chen2022attention, li2022gamma, zheng2022domainplus} for implicit motion alignment, bypassing traditional optical flow techniques. Additionally, some methods integrate optical flow networks \cite{catley2022flexhdr} or deformable convolutions \cite{liu2021adnet} for spatial feature alignment. Non-local attention \cite{yan2020deep} and recurrent networks \cite{prabhakar2021self} have been used for large-scale motion modeling. Along with distinctive merits of Transformer \cite{vaswani2017attention}, \ie, long-range dependencies, which is particularly suitable for capturing large object displacement, recent deghosting approaches based on Transformers \cite{liu2022ghost, song2022selective, yan2023unified, liu2023joint, yan2023smae, tel2023alignment} leverage the traits for large object displacement, with CA-ViT \cite{liu2022ghost} and SCTNet \cite{tel2023alignment} as new baselines using Swin Transformer block \cite{liu2021swin}. However, these methods face challenges: 1) losing Swin Transformer's significantly hierarchical advantage; 2) visible block artifacts in tone-mapped results\textemdash the resulting to partition input images into blocks (\eg, 256$\times $256) for inference, as 2K images cannot be processed on limited-memory GPUs. The latest GDP \cite{fei2023generative} employs a pre-trained denoising diffusion probabilistic model (DDPM) \cite{ho2020denoising} for image restoration, including HDR, but it's limited by less realistic hole filling in occluded areas and slow inference speed.
\vspace{-4mm}
\section{Proposed Method}
\vspace{-3mm}
\label{proposed_method}
\subsection{Preliminaries}
Let $\left\{ {{I_i}} \right\}_{i =1}^n$ be a set of LDR images with varying exposure times $\left\{ {{t_i}} \right\}_{i=1}^n$.
If the input is not in RAW format, they are transformed into the HDR domain via gamma correction, defined by the power-law expansion:
\begin{equation}
\label{eq:eq1}
{H_i} = {{I_i^\gamma } \mathord{\left/
		{\vphantom {{{I_i}^\gamma } {{t_i}}}} \right.
		\kern-\nulldelimiterspace} {{t_i}}},
\end{equation}
where $I_i$ is raised to a gamma value $\gamma >1 $ and divided by time ${t_\emph{i}}$ to get the corresponding image ${H_i}$. In the common case of $\gamma  = 2.2$, inputs and outputs are typically normalized to the interval $\left[ {0,1} \right]$. Similar to previous works \cite{kalantari2017deep, wu2018deep}, we take $\left\{ {{X_i}} \right\}_{i=1}^\emph{n}$ as the input, where ${X_i} = \left[ {{I_i},{H_i}} \right]$ concatenates $I_i$ and $H_i$ along the channel dimension. The acquisition of HDR imaging, \emph{H}, is to find a mapping function $\boldsymbol{\varphi}$ with respect to the parameter $\theta$, satisfying:
\begin{equation}
\label{eq:eq2}
H = \boldsymbol{\varphi} \left( {{X_1},{X_2}, \cdots ,{X_n};\theta } \right).
\end{equation}

In this work, we also set $n=3$ and use ${X_2}$ with a medium exposure as the reference frame, so that the motion content in the resulting HDR image \emph{H} is consistent with it.
\begin{figure}[!t]
\centering
\includegraphics[width=1.0\textwidth,keepaspectratio]{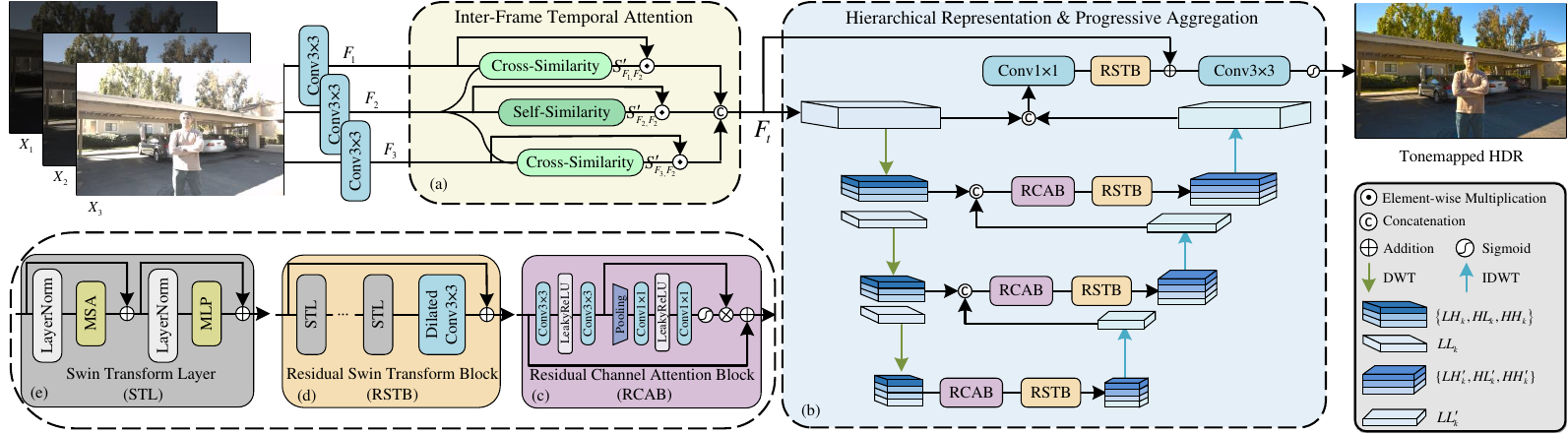}
\vspace{-4mm}
\caption{\textbf{Framework Overview}. The proposed framework mainly consists of three stages, \ie, shallow feature extraction, inter-frame temporal attention, and hierarchical representation \& progressive aggregation.}
\label{fig3}
\vspace{-4mm}
\end{figure}
\vspace{-3mm}
\subsection{Overall Pipeline}
\vspace{-3mm}
Given a set of LDRs $\left\{ {{X_i}} \right\}_{i = 1}^3$, we utilize three separate 3$\times$3 convolutions to extract corresponding shallow features ${F_i} \in \mathbb{R}^{H \times W \times C}$, where ${X_i} \in \mathbb{R}^{H \times W \times 6}$, $H \times W$ denotes the resolution of the image, and \emph{C} denotes the number of channels. The architecture of PASTA, as depicted in \cref{fig3}, comprises two key components, \ie, Inter-Frame Temporal Attention and Hierarchical Representation \& Progressive Aggregation.

\vspace{-4mm}
\subsection{Inter-Frame Temporal Attention}
\label{IFTA}
The exposure stack captures the temporal information across frames. Previous works \cite{yan2019attention, liu2022ghost} utilize spatial attention between reference and non-reference frames to suppress motion and saturation. Different from them, we replace spatial attention with temporal attention \cite{wang2019edvr} because 1) different regions across frames are not equally informative due to exposure variations, occlusions, and moving objects. 2) The spatial attention is implemented by cascading the reference and non-reference frames instead of directly calculating their similarity, and then performing convolution with sigmoid activation to obtain the attention map. In fact, this computation way does not easily discern the differences between frames when both motion and saturation coexist \cite{yan2023unified}. 3) The computation of spatial attention can be regarded as a local form of self-attention mechanism.

IFTA module is illustrated in \cref{fig3}(a) with the aim of computing attention features between adjacent frames in the embedding space. For shallow feature ${F_i} \in {\mathbb{R}^{H \times W\times C}}$, $i \in \left\{ {1,2,3} \right\}$, the similarity matrix $S \in {\mathbb{R}^{H \times W}}$ is obtained by:
\begin{equation}
\label{eq:eq3}
{S_{{F_i},{F_2}}}\left( {x,y} \right) = \left\langle {{\phi _i}\left( {x,y} \right),{\phi _2}\left( {x,y} \right)} \right\rangle,
\end{equation}
where $x$ and $y$ are feature coordinates, ${\phi _i}$ and ${\phi _2}$ are two embeddings of corresponding features ${F_i}$ and ${F_2}$, achieved via a convolution operation, and $\left\langle{\cdot , \cdot } \right\rangle $ denotes the element-wise inner product along the channel dimension. After sigmoid function and channel expansion operations, we can obtain the attention map ${S'_{{F_i},{F_2}}} \in {\mathbb{R}^{H \times W \times C}}$. The contribution of the sigmoid function lies in: i) \textit{highlighting} misaligned and saturated feature regions; ii) By limiting the inner product output to $[0, 1]$, the function \textit{keeps gradients in check} during backpropagation, enhancing \textit{training stability}; iii) improving \textit{convergence efficiency}. The similarity in the \cref{eq:eq3} belongs to an element-wise local attention mechanism. It is noted that \cref{eq:eq3} not only includes cross-similarity but also self-similarity for reference feature. This differs from spatial attention-based methods \cite{yan2019attention, liu2022ghost} which indiscriminately concatenate the reference feature directly into the attention features.
Then, temporal attention features $\left\{ {{F_{i, t}}} \right\}_{i = 1}^3$ can be obtained by element-wise multiplication of ${F_i}$ and ${S_{{F_i},{F_2}}}$, defined as $F_{i, t} = {F_i} \odot {S'_{{F_i},{F_2}}}$.

Lastly, the temporal attention features mentioned above are concatenated by ${F_t} = {\rm{Cat}}\left( {{F_{1,t}},{F_{2,t}},{F_{3,t}}} \right)$.

\vspace{-4mm}
\begin{figure}[!t]
  \centering
  \begin{subfigure}[b]{.185\linewidth}
	\includegraphics[width=\linewidth]{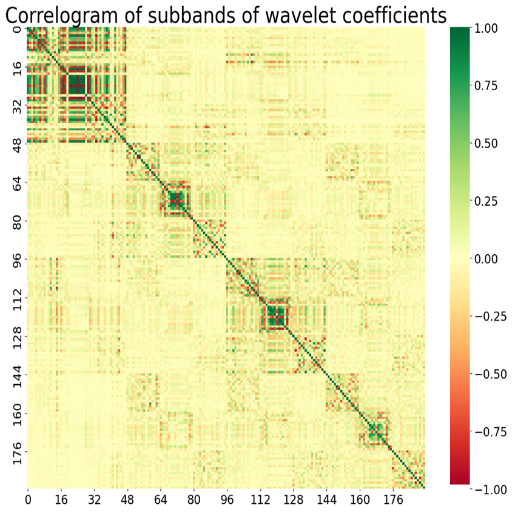}
	\caption{1st level}
    \label{fig:level1}
  \end{subfigure}%
  \hfill
  \begin{subfigure}[b]{.185\linewidth}
	\includegraphics[width=\linewidth]{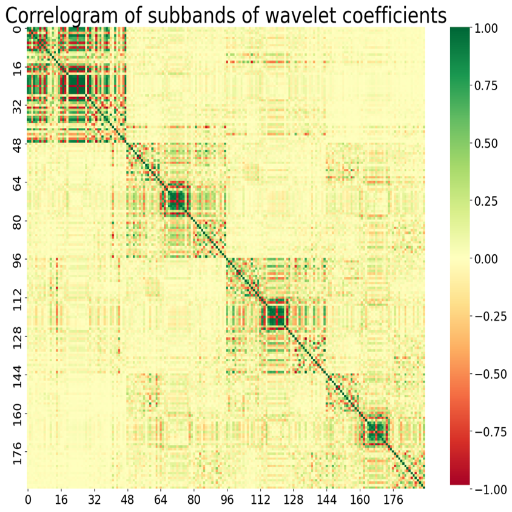}
	\caption{2nd level}
    \label{fig:level2}
  \end{subfigure}%
    \hfill
  \begin{subfigure}[b]{.185\linewidth}
	\includegraphics[width=\linewidth]{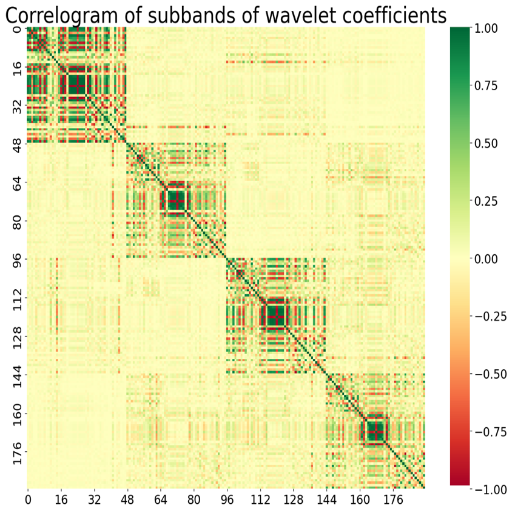}
	\caption{3rd level}
    \label{fig:level3}
  \end{subfigure}%
    \hfill
  \begin{subfigure}[b]{.185\linewidth}
	\includegraphics[width=\linewidth]{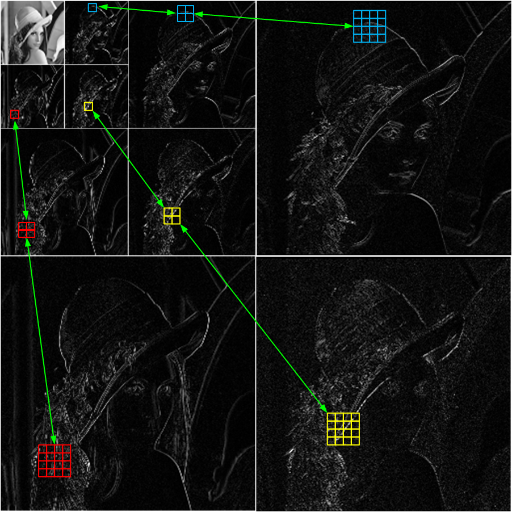}
	\caption{Inter-scale}
    \label{fig:inter}
  \end{subfigure}%
    \hfill
  \begin{subfigure}[b]{.185\linewidth}
	\includegraphics[width=\linewidth]{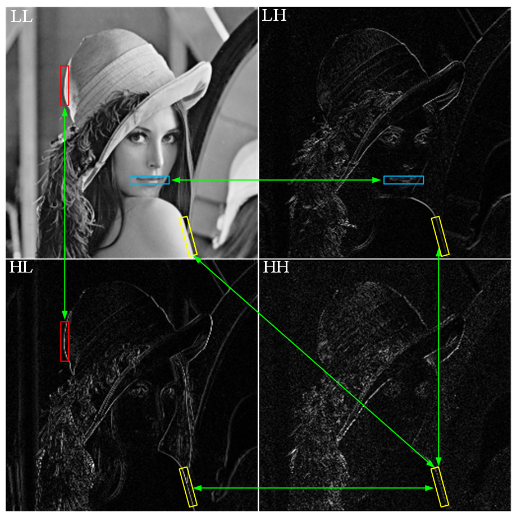}
	\caption{Intra-scale}
    \label{fig:intra}
  \end{subfigure}%
  \caption{\textbf{(a)-(c) Feature Channels Correlation} between subband coefficients of three levels of wavelet decomposition measured by Pearson correlation coefficient (PCCs). The channel dimension of each subband coefficient is 48. In each subfigure, from the top left to the bottom right, the correlations among the channels of the subband coefficients $\left\{ {L{L_k},L{H_k},H{L_k},H{H_k}} \right\}_{k = 1}^3$ themselves are shown. \textbf{(d)-(e) Dependencies between wavelet coefficients} across different scales give rise to a quad-tree structure, where not only the parents but also the children coefficients contain pertinent information.}
  \vspace{-4mm}
  \label{fig:ablation_QUANTIZATION}
\end{figure}

\subsection{Hierarchical Representation \& Progressive Aggregation}
\label{hierarchical_represention}

To reduce the computational complexity of self-attention module, many works utilize pooling \cite{wang2021pyramid}, patch merging \cite{liu2021swin}, and pixel-shuffle \cite{zamir2022restormer} for down-sampling feature maps. These operations nevertheless are either irreversible \cite{wang2021pyramid, liu2021swin}, which potentially lead to a drop in the capability of feature representation, especially for low-level pixel-wise visual tasks, or still suffer from excessive memory in case of high-resolution images due to the symmetric U-Net architecture \cite{zamir2022restormer}. Among of them, Swin \cite{liu2021swin} has demonstrated that hierarchy is of vital importance in vision tasks. In addition to the parallel computation achieved by shift window, the superiority of Swin over ViT \cite{dosovitskiy2020image} is mainly attributed to this paramount trait. However, existing Transformer-based deghosting works \cite{liu2022ghost, liu2023joint, yan2023unified, yan2023smae}, as well as SwinIR \cite{liang2021swinir} represented by the use of Swin blocks, overlook or do not fully inherit the character of hierarchical representation. 

The attractive properties of the wavelet transform enable it to well represent and decouple features, including \textbf{\textit{locality}}, \textbf{\textit{multi-resolution}}, \textbf{\textit{compression}}, \textbf{\textit{clustering}} and \textbf{\textit{persistence}} \cite{crouse1998wavelet,romberg2001bayesian}. \cref{fig:inter,fig:intra} show the magnitude values of wavelet coefficients at three scales and indicate that subbands are not statistically independent from visual inspection, Due to the non-Gaussianity of wavelet marginals, previous works have primarily utilized hidden Markov models \cite{romberg2001bayesian,crouse1998wavelet} to characterize the statistical dependencies between those coefficients. In this work, our objective is to develop a module to implicitly explore the relationships among subbands, which is rich and flexible enough to represent coefficient characteristics yet concise, tractable and efficient for high-resolution images. 

\vspace{-4mm}
\subsubsection{Wavelet Representation.}
DWT, as a classic technique for multi-resolution, decomposes a 2D image into four subband components: one low-frequency band, \textit{LL}, and three high-frequency ones, \ie, \textit{LH}, \textit{HL}, and \textit{HH}, with half the resolution of the source. The DWT is reversible, and its inverse, denoted as IDWT. As shown in \cref{fig3}(b), let $L{L_0} = {F_t}$ and we recursively perform the DWT on the low-frequency component $\left\{ {L{L_k}} \right\}_{k= 0}^{K-1}$ at each scale \emph{k}, obtaining a set of a set of high-frequency residual components $\left\{ {L{H_k},\emph{H}{L_k},H{H_k}} \right\}_{k = 1}^K$ and the coarset low-frequency approximated one $L{L_K}$. The low-frequency features identify the motion between frames, illumination variations, and color information, while the high-frequency ones capture contour information and rapid changes in appearance. As the decomposition level \emph{k} increases, the resolution of the corresponding high- and low-frequency components decreases from $H \times W$ to $\frac{H}{{{2^k}}} \times \frac{W}{{{2^k}}}$ while keeping the channel dimension unchanged. Compared to the U-Net architecture of Restormer \cite{zamir2022restormer}, our representation structure has no additional learnable parameters except for the fixed ones of the wavelet filters in terms of hierarchical representation. Furthermore, the channel dimension of Restormer doubles as the decomposition levels increase.

\vspace{-4mm}
\subsubsection{Coarse-to-Fine Aggregation.}
The wavelet representation not only expands the receptive field for the aggregation stage, which will foster ghost removal for large object displacement and large-disparity alignment, but also achieves feature decoupling, specifically for low- and high-frequency components. Recalling the remarkable properties of wavelet coefficients and the dependency on subband coefficients stated in \cref{hierarchical_represention}, this inspires us to implicitly explore and utilize the statistical relationship.
\cref{fig:level1,fig:level2,fig:level3} illustrates the correlation coefficients between different frequency subband channels measured by PCCs.  
Correlation and redundancy among wavelet subband channels are evident. 
As the levels of DWT decomposition increase, the inter-channel correlation gradually intensifies.

As shown in \cref{fig3}(b), after three levels of DWT, we obtain the final low-frequency subband $L{L_3}$ and the corresponding detail ones $\left\{ {L{H_3},H{L_3},H{H_3}} \right\}$. Taking into account the correlation between different frequency bands and the redundancy of subband features, we process them by concatenating the high and low-frequency subbands after wavelet decomposition. Another purpose of joint processing is to prevent the loss of spatial coherence between low and high frequencies when they are processed separately. It is common for the low frequencies to remove ghosting, but the high frequencies fail to handle the boundaries or contours of moving objects effectively, thereby resulting in ghosting. The developed progressive aggregation strategy for large-scale dynamic scenes mainly includes four stages. The core module is composed of a residual channel attention block (RCAB) \cite{zhang2018image} and a residual Swin Transformer block (RSTB) \cite{liang2021swinir}, where RCAB is used to explore inter-subband dependencies, followed by RSTB to build long-range spatial dependencies. The first three stages involve aggregating within the space of wavelet representation with spatial contextual information. From bottom to top, \ie, from global to local, our progressive aggregation framework iteratively upscales and refines the estimates from the previous level by predicting the high- and low-frequency coefficients. 
The last stage operates in the original feature space (without downsampling operations), thereby preserving fine details in the final reconstructed HDR image. Specifically, the reconstructed features $L{{L'}_0}$ are concatenated with the original features $F_t$. Then a 1$\times$1 convolution is used to achieve channel dimension compression, followed by passing through an RSTB module.

\noindent \textbf{Skip Connection.} To construct a ghost-free HDR image, we have developed a model with a long skip connection extending from the IFTA feature $F_t$. This connection refines spatio-temporal aggregated features by merging the reconstructed wavelet coefficient features from the initial three stages with the finest features of the last stage. We have named this model \textbf{PASTA-I}, highlighting its capability for implicit alignment, which significantly reduces motion blur and ghosting effects in case of shooting moving scenes.

\vspace{-4mm}
\subsubsection{w and w/o IFTA Module.} As the proposed model \textbf{PASTA-I} without IFTA strategy will simplify to cascading shallow features directly, akin to \cite{wu2018deep, niu2021hdr}, this subsection emphasizes the comparison of the developed variant with them. As we previously discussed, 1) Both methods \cite{wu2018deep,niu2021hdr} are vulnerable to the cases with occlusions and large moving objects. 2) Self-attention inherently possesses alignment capabilities to some extent. In contrast, applying the self-attention mechanism to \cite{wu2018deep, niu2021hdr}, \ie, our alignment-free variant \textbf{PASTA-U}, is competent to handle these scenarios. To circumvent the generation of ghosting artifacts, \textbf{PASTA-I}'s long skip ${F_t}$ is replaced in PASTA-U with ${{\mathop{\rm Conv}\nolimits} _{3 \times 3}}\left( {{F_2}} \right)$, which considers the motion in intermediate frames to direct network learning. The results from PASTA-U show that: i) Self-attention can directly leverage unaligned feature information, and ii) A hierarchical structure can facilitate the enhancement of the self-attention mechanism's ability to utilize unaligned feature information. This is actually evidenced by the fact that Swin \cite{liu2021swin} outperforms ViT \cite{dosovitskiy2020image}.

\vspace{-3.5mm}
\subsection{Loss Function}

We use $\mu$-law function to map the HDR image from the linear domain to the tonemapped domain through $\mathcal{T}\left( x \right) = \frac{{\log \left( {1 + \mu x} \right)}}{{\log \left( {1 + \mu } \right)}}$,

where $\mathcal{T}\left(  \cdot  \right)$ is the tone mappting function, $\mu=5000$. In the tonemapped domain, in addition to the previous loss functions \cite{liu2022ghost, yan2023unified} that consist of $\mathcal{L}_1$ loss and perceptual loss ${\mathcal{L}_p}$, we further incorporate a edge loss ${\mathcal{L}_e}$ to surpass the edge blurring possibly caused by the hierarchical representation, which is defined as:

\begin{equation}
\label{eq:eq9}
{\mathcal{L}_\emph{e}} = \sqrt {{{\left\| {\nabla \left( {\mathcal{T}(H)} \right) - \nabla \left( {\mathcal{T}(\hat{H})} \right)} \right\|}^2} + {\varepsilon ^2}},
\end{equation}
where \emph{$\hat{H}$} is the estimated HDR image, and the constant $\varepsilon$ is empirically set to ${10^{ - 3}}$. 

Overall, our loss function is $\mathcal{L} = {\mathcal{L}_1} + \alpha {\mathcal{L}_p} + \beta {\mathcal{L}_e}$, where~$\alpha$~and~$\beta$~are set to 0.01 and 1, respectively.

\vspace{-4mm}
\section{Experiments}
\vspace{-3mm}
\subsection{Experimental Settings}
\noindent{\bf Datasets.} We train the proposed model on Kalantari~\etal~'s \cite{kalantari2017deep} and Tel~\etal~'s~\cite{tel2023alignment} datasets, which consist of 74/15 training/testing scenes and 108/36 counterparts, respectively.

We first crop the images in the training set into image patches of size 128$\times$128 with a stride of 64, and then perform a clockwise rotation of 90$^\circ$, and horizaontal and vertical flips for data augmentation. To validate the effectiveness of the proposed method, we also conduct testing on Sen \etal's \cite{sen2012robust} and Tursen \etal's \cite{tursun2015state} datasets. 

\noindent{\bf Metrics.} We use five commonly used metrics, \ie, PSNR-\emph{l}, PSNR-$\mu$, SSIM-\emph{l}, SSIM-$\mu$, and HDR-VDP-2 \cite{mantiuk2011hdr}, to evaluate the performance of different methods. Here, `-\emph{l}' and `-$\mu$' refer to the calculation of the respective metrics in the linear and tonemapped domain, respectively. HDR-VDP-2 is a visual metric tailored for HDR and its two parameters are set to 25 and 0.5, respectively, for the Kalantari \etal's dataset. To ensure a fair comparison, we follow the same parameter settings \cite{tel2023alignment} on Tel \etal's dataset.

\noindent{\bf Implementation Details.} Our model is implemented based on PyTorch and uses the Adam optimizer with an initial learning rate of 2$e$-4, where ${\beta _1}$, ${\beta _2}$, and $\varepsilon $ are set to 0.9, 0.999, and 1$e$-8, respectively. The total number of iterations is 300K, and the learning rate is halved every 50K ones. We set the batch size to 16 and trained on 8 NVIDIA TITAN Xp GPUs. 
\cref{tab:model_parameter_details} lists the parameter configurations, highlighting the tiny version's key differences in channel dimension, Swin Transformer layers (STL), and multi-head self-attention heads.

\begin{table}[th]
  \centering
  \vspace{-5mm}
  \scriptsize
    \caption{Model architecture configuration. ``\emph{C}'', ``Reduction'' and ``Win. sz.'' refer to the channel dimension in shallow feature ${F_i}$, the channel reduction factor for each RCAB, and the window size in the RSTB, respectively.}
    \vspace{-2.5mm}
    \resizebox{\linewidth}{!}{
    \begin{tabular}{c|c|c|c|c|c|c|c}
    \hline
    Model & \emph{C} & Num. of STL & Num. of heads & Reduction & MLP ratio & Win. sz. & \#Param.\\
    \hline
    PASTA-I & 16 & [6, 6, 6, 6] & [4, 4, 4, 4] & [4, 4, 4] & 2 & 8 & 8.163M\\
    PASTA-I-Tiny & 12 & [2, 2, 2, 2] & [3, 3, 3, 3] & [4, 4, 4] & 2 & 8 & 2.636M\\
    PASTA-U & 16 & [6, 6, 6, 6] & [4, 4, 4, 4] & [4, 4, 4] & 2 & 8 & 8.165M \\
    PASTA-U-Tiny & 12 & [2, 2, 2, 2] & [3, 3, 3, 3] & [4, 4, 4] & 2 & 8 & 2.637M\\
    \hline
    \end{tabular}%
    }
  \label{tab:model_parameter_details}%
  \vspace{-5.5mm}
\end{table}%
\vspace{-4mm}
\subsection{Comparison with SOTAs}
\vspace{-2mm}
We evaluate PASTA against 16 SOTA methods that include Sen \etal \cite{sen2012robust}, Hu \etal \cite{hu2013hdr}, Kalantari \etal \cite{kalantari2017deep}, DeepHDR \cite{wu2018deep}, AHDRNet \cite{yan2019attention}, PAMNet \cite{pu2020robust}, NHDRRnet \cite{yan2020deep}, PSFNet \cite{ye2021progressive}, FSHDR \cite{prabhakar2021labeled}, HDRRNN \cite{prabhakar2021self}, HDR-GAN \cite{niu2021hdr}, APNT \cite{chen2022attention}, FHDRNet \cite{dai2024wavelet}, STHDR \cite{song2022selective}, CA-ViT \cite{liu2022ghost}, and SCTNet \cite{tel2023alignment}.
\begin{table}[!t]
\centering
\scriptsize
\caption{Quantitative comparison with state-of-the-art methods on Kalantari \etal's \cite{kalantari2017deep} and Tel \etal's \cite{tel2023alignment} datasets. ``-'' indicates that the code or pretrained weight is not available, and the parameters for calculating HDR-VDP-2 (H.V.-2) are unspecified. The top two performers under each metric are highlighted in \textcolor{red}{red} and \textcolor{blue}{blue}, respectively.}
\vspace{-2.5mm}
\resizebox{\linewidth}{!}{
	\begin{tabular}{clcccccc}
		\toprule[1.25pt]
		\multicolumn{1}{c}{\multirow{2}[2]{*}{Dataset}} &
        \multicolumn{1}{l}{\multirow{2}[2]{*}{Method}} &
        \multicolumn{1}{c}{\multirow{2}[2]{*}{Venue}} &
		\multicolumn{1}{c}{\multirow{2}[2]{*}{H.V.-2 $\uparrow$}} &
		\multicolumn{2}{c}{Linear Imgs.} & \multicolumn{2}{c}{Tonemapped Imgs.} \\
		\cmidrule(r){5-6} \cmidrule(r){7-8}
		& & & & \multicolumn{1}{l}{PSNR-\emph{l}$\uparrow$} & \multicolumn{1}{l}{SSIM-\emph{l}$\uparrow$} & \multicolumn{1}{l}{PSNR-$\mu$$\uparrow$} & \multicolumn{1}{l}{SSIM-$\mu$$\uparrow$} \\
  
		\midrule
		\multirow{16}{*}{Kalantari} & Sen \etal \cite{sen2012robust} & TOG'12  &  59.38  & 38.11  &  0.9721  & 40.80  & 0.9808   \\
		& Hu \etal \cite{hu2013hdr} & CVPR'13 & - & 30.76  & 0.9503 & 35.79 & 0.9717   \\
		& Kalantari \etal \cite{kalantari2017deep} & TOG'17 & 64.18 & 41.23 & 0.9846  & 42.67  & 0.9888   \\
		& DeepHDR \cite{wu2018deep} & ECCV'18& 64.90 & 40.88 & 0.9858 & 41.65  & 0.9860  \\
		& AHDRNet \cite{yan2019attention} &CVPR'19 & 65.61 & 41.14  & 0.9702 & 43.63  & 0.9900  \\
		& PAMNet \cite{pu2020robust} & ACCV'20& - & 41.65 & 0.9870 & 43.85  & 0.9906   \\
		& NHDRRnet \cite{yan2020deep} &TIP'20 & 64.08 & 41.08 & 0.9861 & 42.41 &0.9887 \\
		& PSFNet \cite{ye2021progressive} &MM'21 &  - & 41.57 & 0.9867 & 44.06  & 0.9907 \\
		& HDRRNN \cite{prabhakar2021self} & TCI'21& - & 41.68 & - & 42.07  & -  \\
		& HDR-GAN \cite{niu2021hdr}&TIP'21 & 65.54 & 41.57 & 0.9865 & 43.92  & 0.9905  \\
		& APNT \cite{chen2022attention}& TIP'22 & - & 41.61 & 0.9879 & 43.94 & 0.9898  \\
        & FHDRNet \cite{dai2024wavelet} & CVIU'24 & 65.71 & 41.47 & 0.9869 & 43.91 & 0.9908\\
		& STHDR \cite{song2022selective}& ECCV'22 & - & 41.70 & 0.9872 & 44.10 & 0.9909 \\
		& CA-ViT \cite{liu2022ghost}& ECCV'22 & 65.66 & 42.18 & 0.9884 & 44.32  & 0.9916 \\
        & SCTNet \cite{tel2023alignment}& ICCV'23 & 64.48 & 42.29 &0.9887 & \textbf{\textcolor{blue}{44.49}} & \textbf{\textcolor{red}{0.9924}} \\
		& \textbf{PASTA-I}& Ours & \textbf{\textcolor{red}{65.92}} & \textbf{\textcolor{red}{42.50}} & \textbf{\textcolor{red}{0.9899}} & \textbf{\textcolor{red}{44.53}} & \textcolor{blue}{\textbf{0.9918}}  \\
		& \textbf{PASTA-U} & Ours & \textbf{\textcolor{blue}{65.86}} & \textbf{\textcolor{blue}{42.45}} & \textbf{\textcolor{blue}{0.9895}} & \textbf{\textcolor{blue}{44.49}} & 0.9917 \\

        \midrule
		\multirow{8}{*}{Tel} & Kalantari \etal \cite{kalantari2017deep} & TOG'17 & 67.09 & 43.37 & 0.9924  & 40.05  & 0.9794   \\
		& NHDRRnet \cite{yan2020deep} & TIP'20 & 65.41 & 39.61 & 0.9853 & 36.68 &0.9590 \\
		& AHDRNet \cite{yan2019attention} & CVPR'19 & 68.80 & 45.30  & 0.9943 & 42.08  & 0.9837  \\
        & FHDRNet \cite{dai2024wavelet} & CVIU'24 & 69.86 & 45.80 & 0.9948 & 42.41 & 0.9858\\
		& CA-ViT \cite{liu2022ghost} & ECCV'22 & 69.23 & 46.35 & 0.9948 & 42.39 & 0.9844 \\
		& SCTNet \cite{tel2023alignment} & ICCV'23 & 70.66 & 47.51 & \textbf{\textcolor{blue}{0.9952}} & 42.55 & 0.9850\\
		& \textbf{PASTA-I} & Ours & \textbf{\textcolor{red}{71.30}} & \textbf{\textcolor{red}{48.24}} & \textbf{\textcolor{red}{0.9961}} & \textbf{\textcolor{blue}{43.53}} & \textcolor{blue}{\textbf{0.9883}}  \\
		& \textbf{PASTA-U} & Ours & \textbf{\textcolor{blue}{71.20}} & \textbf{\textcolor{blue}{48.16}} & \textbf{\textcolor{red}{0.9961}} & \textbf{\textcolor{red}{43.62}} & \textbf{\textcolor{red}{0.9884}}\\
		\bottomrule[1.25pt]
\end{tabular}}
\label{table:sig17_iccv23}%
\vspace{-4.0mm}
\end{table}

\noindent{\bf Quantitative Comparison.} \cref{table:sig17_iccv23} shows that PASTA-I and PASTA-U perform competitively on two benchmarks, excelling particularly on the latest \cite{tel2023alignment}. Notably, both models significantly outperform the latest baseline, CA-ViT, by margins of 0.32dB/0.26dB and 0.21dB/0.17dB in PSNR-\emph{l}/PSNR-$\mu$ on Kalantari~\etal's dataset. Additionally, we have observed that HDR-VDP-2 metrics do not consistently exhibit proportionality to PSNR values. On Tel \etal's dataset, our methods surpass top contenders like SCTNet and CA-ViT across all metrics. PASTA-U's performance suggests that self-attention allows for effective ghost-free HDR image reconstruction even with some large misalignment. It should be noted that some of the values in \cref{table:sig17_iccv23} are derived from \cite{yan2023unified,liu2022ghost,tel2023alignment}.

\begin{figure}[!t]
\centering
\begin{overpic}[width=0.9\linewidth,tics=2]{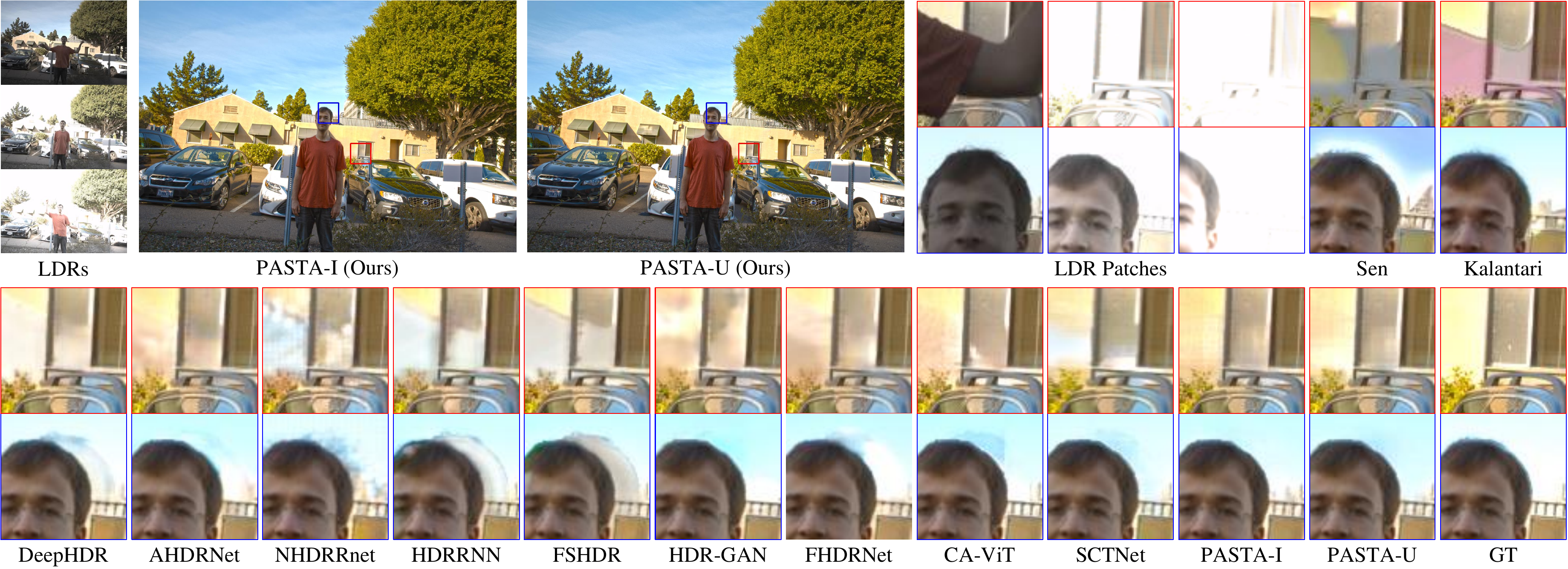}
\end{overpic}
\vspace{-4mm}
\caption{Visual comparison of large-scale foreground and dense background motion with SOTA methods on Kalantari \etal's dataset \cite{kalantari2017deep}.}
\label{fig:sig17}
\end{figure}

\begin{figure*}[!t]
\centering
\begin{overpic}[width=0.95\linewidth,tics=2]{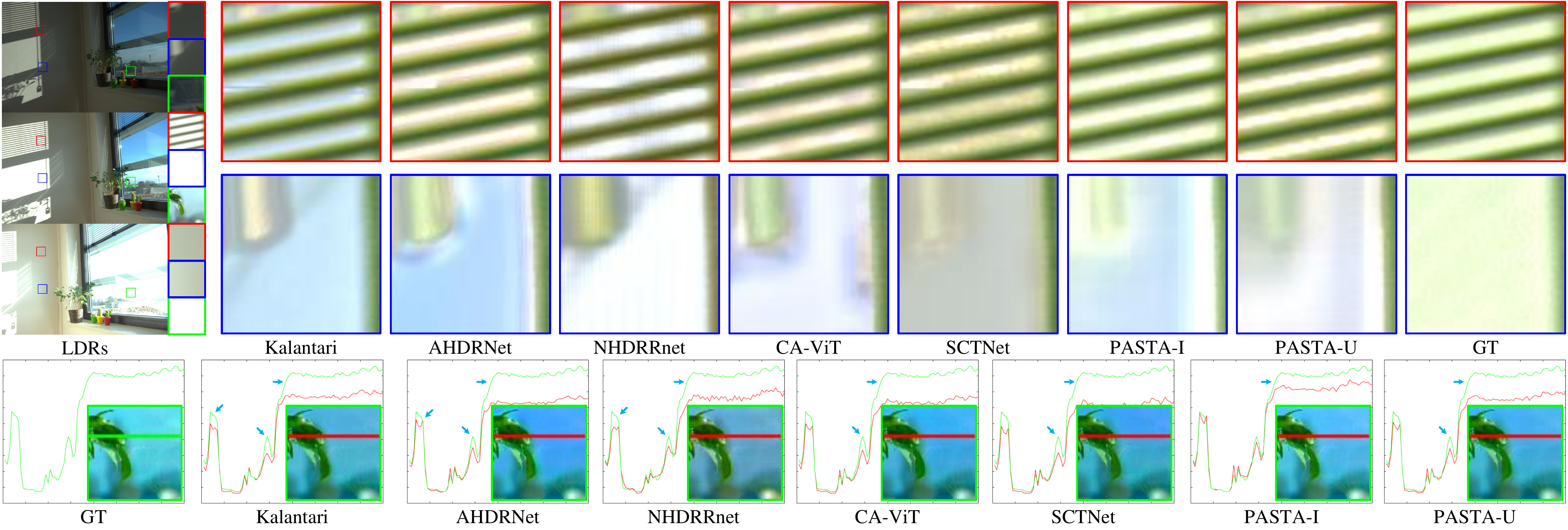}

\end{overpic}
\vspace{-2mm}
\caption{Visual comparison of large disparities with SOTA methods on Tel \etal's dataset \cite{tel2023alignment}. The bottom row displays 1D intensity shift of the green close-up regions labeled with the line. Zoom in for better viewing.}
\label{fig:iccv23}
\vspace{-3mm}
\end{figure*}

\noindent{\bf Qualitative Comparison.}
\cref{fig:sig17,fig:iccv23} show two challenging scenes: large-scale foreground motion and significant camera motion with large disparities. The coexistence of occlusions and saturation is inevitable and intractable in the presence of exposure and motion variations. In such case, non-attention or non-transformer-based methods like the approaches of Sen \etal and Kalantari \etal, DeepHDR, NHDRRnet, HDRRNN and FSHDR struggle with ghosting artifacts due to unreliable alignment (\eg, optical flow, homography) or lack of feature alignment, often failing in fine detail preservation, such as hair in \cref{fig:sig17}. Additionally, the patch-based method of Sen \etal incurs noticeable halos. In contrast, our method surpasses others like CA-ViT, SCT-Net and AHDRNet in ghosting removal, saturation handling, detail retention, and color consistency. In window scenes with significant disparities, approaches like Kalantari \etal and NHDRRnet suffer content loss in saturated areas, while CA-ViT and SCTNet struggle with fine detail recovery and noise suppression, as shown in \cref{fig:iccv23}'s red close-up regions. By comparison, our method achieves relatively noise-free results in these areas. The proposed method, especially PASTA-I, also preserve edge sharpness and intensity in smooth regions better than competitors like CA-ViT and SCTNet, evident in the green close-ups in \cref{fig:iccv23}. For additional results, please refer to the supplementary materials.

\begin{table*}[!t]
    \caption{Ablation study including IFTA, RCAB, ${\mathcal{L}_p}$ and ${\mathcal{L}_e}$ on the Kalantari \etal’ dataset \cite{kalantari2017deep} using PASTA-I ({\small{Ours}}). ``SA vs. IFTA'' refers to IFTA being replaced with spatial attention.}
  \centering
  \scriptsize
  \setlength{\tabcolsep}{3mm}
  \begin{tabular}{cccccccc}
      \toprule
      Metrics & \makecell{SA vs.\\ IFTA} & \makecell{w/o\\IFTA} & \makecell{w/o\\ RCAB} & \makecell{w/o \\${\mathcal{L}_p}$} & \makecell{w/o \\ ${\mathcal{L}_e}$} &  PASTA-U & PASTA-I \\
      \midrule
      PSNR-\emph{l} & 41.86 & 42.33 & 42.19 & 42.61 & 42.45 & 42.45 & 42.50 \\
      SSIM-\emph{l} & 0.9897 & 0.9896 & 0.9895 & 0.9897 & 0.9893 & 0.9895 & 0.9899 \\
      PSNR-$\mu$ & 44.40 & 44.45 & 43.96 & 44.46 & 44.40 & 44.49 & 44.53 \\
      SSIM-$\mu$ & 0.9918 & 0.9916 & 0.9917 & 0.9919 & 0.9917 & 0.9917 & 0.9918 \\
      \bottomrule
  \end{tabular}
  \label{tab:partition_ablation}
\end{table*}
\vspace{-4mm}
\subsection{Ablation Study}
\vspace{-2mm}

\begin{table}[th]
\centering
\caption{Ablation studies of sampling methods and DWT scales. ``PUS vs DWT'' indicates using PixelUnshuffle (PUS) \cite{zamir2022restormer} to replace DWT for downsampling, and Pixelshuffle to replace IDWT for upsampling. ``Conv. vs DWT'' denotes the substitution of the DWT with convolution, as used in \cite{wu2018deep,yan2020deep,dai2024wavelet}. ``$K$'' is the scale of the hierarchical representation structure.}
\vspace{-2.5mm}
\resizebox{\linewidth}{!}{
	\scriptsize
	\setlength{\tabcolsep}{2mm} 
	\begin{tabular}{clcccccc}
	\toprule[1.25pt]
	\multicolumn{1}{c}{\multirow{2}[2]{*}{Settings}} & \multirow{2}[2]{*}{Variant} & \multicolumn{2}{c}{Kalantari~\etal's \cite{kalantari2017deep}} & \multicolumn{2}{c}{1000$\times$1500}  &   \multicolumn{2}{c}{2560$\times$1440} \\
	\cmidrule(lr){3-4}\cmidrule(lr){5-6}\cmidrule(lr){7-8}
	& & \#Imgs-\emph{l} & \#Imgs-$\mu$ & \hspace{.3em} G. Mem.  & \hspace{-.5em} Time   & \hspace{.3em} G. Mem.     & \hspace{-.8em}Time\\
	\midrule
    \multirow{2}{*}{Sampling methods} & PUS vs. DWT & 37.19/.9775 & 40.23/.9852 & 7.729 & 2.353 & OOM & -\\
    &Conv. vs. DWT & 42.07/.9880 & 44.52/.9916  & 7.725 & 2.516 & 18.399 & 5.871 \\
    \cdashline{1-8}[1pt/1pt]
    \multirow{3}{*}{DWT scales} & PASTA-I ($K$=1) & 42.20/.9887 & 44.46/.9917 & 7.298 & 2.382  &  17.464 &  5.520 \\
	&PASTA-I ($K$=2) & 42.39/.9893 & 44.42/.9917 & 7.629 & 2.511 & 17.968 & 5.639 \\
	&PASTA-I ($K$=4) & 42.31/.9895 & 44.35/.9916 & 7.869 & 2.550 & 19.584 & 5.942 \\
    \cdashline{1-8}[1pt/1pt] 
	\multirow{2}{*}{\makecell{Our plain \\version with $K$=3}}& PASTA-I & 42.50/.9899 & 44.53/.9918 & 7.850 & 2.543 & 18.743 & 5.891 \\
	&PASTA-I-Tiny & 42.25/.9890 & 44.46/.9917 & 6.015 & 0.829 & 14.399 & 1.889 \\

	\bottomrule[1.25pt]
	\end{tabular}
 }
	\label{tab:downsample_ways_scale}
    \vspace{-4mm}
\end{table}

\begin{table}[th]
\centering
\caption{Comparison of GPU memory (G) consumption and average inference time (seconds) for three LDR images with different resolutions, \ie, 1000$\times$1500 and 2560$\times$1440 (2K), implements on a RTX 3090 GPU (24G). OOM and `-’ refer to the out-of-memory and being unavailable, respectively.}
\vspace{-2.5mm}
\resizebox{\linewidth}{!}{
	\scriptsize
	\setlength{\tabcolsep}{2mm} 
	\begin{tabular}{lcccccccc}
	\toprule[1.25pt]
	   \multirow{2}[2]{*}{Method} & \multicolumn{2}{c}{Tel~\etal's \cite{tel2023alignment}} & \multicolumn{2}{c}{Kalantari~\etal's \cite{kalantari2017deep}} & \multicolumn{2}{c}{1000$\times$1500}  &   \multicolumn{2}{c}{2560$\times$1440} \\
		\cmidrule(lr){2-3}\cmidrule(lr){4-5}\cmidrule(lr){6-7}\cmidrule(lr){8-9}
		 &\#Imgs-\emph{l} & \#Imgs-$\mu$ &\#Imgs-\emph{l} & \#Imgs-$\mu$ & \hspace{.3em} G. Mem.  & \hspace{-.5em} Time   & \hspace{.3em} G. Mem.     & \hspace{-.8em}Time\\
		\midrule
		AHDRNet \cite{yan2019attention} {(CVPR'19)} & 45.30/.9943 & 42.08/.9837 & 41.14/.9702 & 43.63/.9900 & 9.069 & 0.348 & OOM & - \\
        FHDRNet \cite{dai2024wavelet} {(CVIU'24)} & 45.80/.9948 & 42.41/.9858 & 41.47/.9869 & 43.91/.9908 & 5.643 & 0.329 & 13.449 & 0.985 \\
        \cdashline{1-9}[1pt/1pt]
		CA-ViT \cite{liu2022ghost} {(ECCV'22)} & 46.35/.9948 & 42.39/.9844 
 & 42.18/.9884 & 44.32/.9916 & 10.637 & 4.631 & OOM & - \\
		SCTNet \cite{tel2023alignment} {(ICCV'23)} & 47.51/.9952 & 42.55/.9850 & 42.29/.9887 & 44.49/.9924 & 9.732 & 6.621 & OOM & - \\
        \cdashline{1-9}[1pt/1pt]
		PASTA-I & 48.24/.9961 & 43.53/.9883 & 42.50/.9899 & 44.53/.9918 & 7.850 & 2.418 & 18.743 & 5.969 \\
		PASTA-I-Tiny & 48.08/.9957 & 43.45/.9878 & 42.25/.9890 & 44.46/.9917 & 6.015 & 0.867 & 14.399 & 2.071 \\
        \cdashline{1-9}[1pt/1pt]
		PASTA-U & 48.16/.9961 & 43.62/.9884 & 42.45/.9895 & 44.49/.9917 & 7.850 & 2.242 & 18.743 & 5.765 \\
		PASTA-U-Tiny & 48.08/.9960 & 43.51/.9879 & 42.07/.9883 & 44.43/.9916 & 6.015 & 0.773 & 14.402 & 2.036 \\
		\bottomrule[1.25pt]
	\end{tabular}
 }
\label{tab:efficiency_analysis}
\vspace{-4mm}
\end{table}

Here, taking \textbf{PASTA-I} as an example, we validate our framework's efficacy via IFTA, RCAB, edge loss, among others, as detailed in~\cref{tab:partition_ablation,tab:downsample_ways_scale}.

\noindent{\bf Effectiveness of IFTA.} To validate the effectiveness of IFTA, we replaced it with spatial attention\cite{yan2019attention, liu2022ghost}, \ie, SA vs. IFTA, and directly removed it, \ie, w/o IFTA, respectively. The performance of the variant based on spatial attention is slightly lower than w/o IFTA and our full model, especially in terms of PSNR-\emph{l} by a margin of 0.47dB and 0.64dB, respectively. As we mentioned in \cref{IFTA}, i) The indiscriminate concatenation of reference frame features into attention features by spatial attention may degrade the network performance in case of occlusions and saturation coexisting in the reference frames; ii) The computation of spatial attention can be seen as a localized form of self-attention mechanism, which also indicates that self-attention has a certain alignment effect.

\noindent{\bf Effectiveness of RCAB.}
The RCAB module is integral to our framework, focusing on the dependencies both within and across scales of wavelet subband coefficients. When RCAB is removed, we observe a significant reduction in both PSNR-\emph{l} and PSNR-$\mu$ by 0.31dB and 0.57dB. This indicates that the superiority of our framework lies in its implicit statistical modeling of wavelet coefficients.

\noindent{\bf Effectiveness of Edge Loss.} The variant w/o ${\mathcal{L}_p}$, \ie., only $\mathcal{L}_1$ and ${\mathcal{L}_e}$, achieved a higher PSNR-\emph{l} of 42.61dB, suggesting that this loss prevents edge blurring during the process of progressive aggregation, thereby reducing or avoiding halo artifacts. Additionally, its SSIM-$\mu$ slightly improved to 0.9919, indicating a potential trade-off between the objective quality of exposure and the perceived quality of texture and structure. The variant, w/o ${\mathcal{L}_e}$, falls slightly behind in all metrics compared to the variant w/o ${\mathcal{L}_p}$. Our full model indicates that ${\mathcal{L}_e}$ plays a subtle yet tangible role in reconstructing high-frequency detail, while edge loss in preserving edge sharpness. 

\noindent{\bf Sampling methods and DWT scales.} We also replace the DWT operation with alternative sampling methods. The results, listed in the upper section of \cref{tab:downsample_ways_scale}, affirm our efficacy. In comparison with Conv-based methods,
our PASTA-I-tiny achieved a +0.18dB improvement with only 77\% memory utilization, simultaneously accelerating inference time by a factor of $\times$3.

\subsection{Efficiency Analysis}

\noindent{\bf Wavelet Representation.} Efficiency is achieved through the use of DWT for feature extraction. DWT requires no additional learned parameters and isotropic feature extraction avoids increasing channel capacity. It efficiently captures multi-scale information.

\noindent{\bf Progressive Hierarchical Aggregation.} Our progressive hierarchical aggregation strategy speeds up the network by processing lower-resolution representations and iteratively refining them. This approach efficiently retains essential information while discarding redundant details.

\noindent{\bf Quantitative Results.} \cref{tab:efficiency_analysis} reports GPU memory consumption (G. Mem.)\footnote{Computed by PyTorch {torch.cuda.max\_memory\_allocated()} function.} and average inference time of our methods and four baselines on three LDR inputs of different resolutions, calculated over 100 iterations. Taking the resolution of 1000$\times$1500 as an example, benefiting from the hierarchical representation and progressive aggregation, the proposed method achieves GPU costs of $74\%$ compared to CA-ViT and $80\%$ to SCTNet; the inference speed is nearly 2.8 times faster than SCTNet, 2 times than CA-ViT. When inputs are 2K, aside from the FHDRNet, none of the aforementioned baselines can directly perform inference.~\cref{table:sig17_iccv23,tab:efficiency_analysis} indicate that the proposed method achieves the best balance in inference time, GPU memory usage and reconstruction performance compared to SOTA methods, especially those \cite{liu2022ghost,tel2023alignment,yan2023unified} utilizing Transformer.

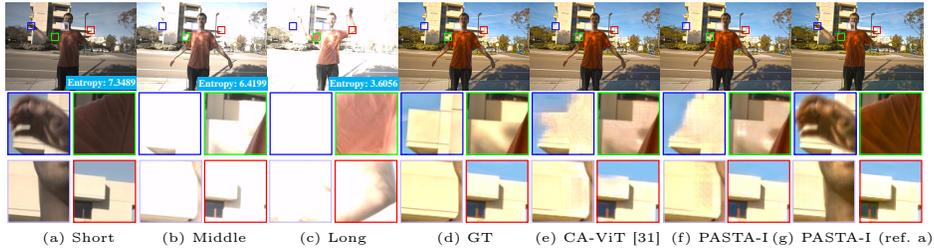
\begin{figure*}[t]
	\centering
    \begin{overpic}[width=1.0\textwidth]{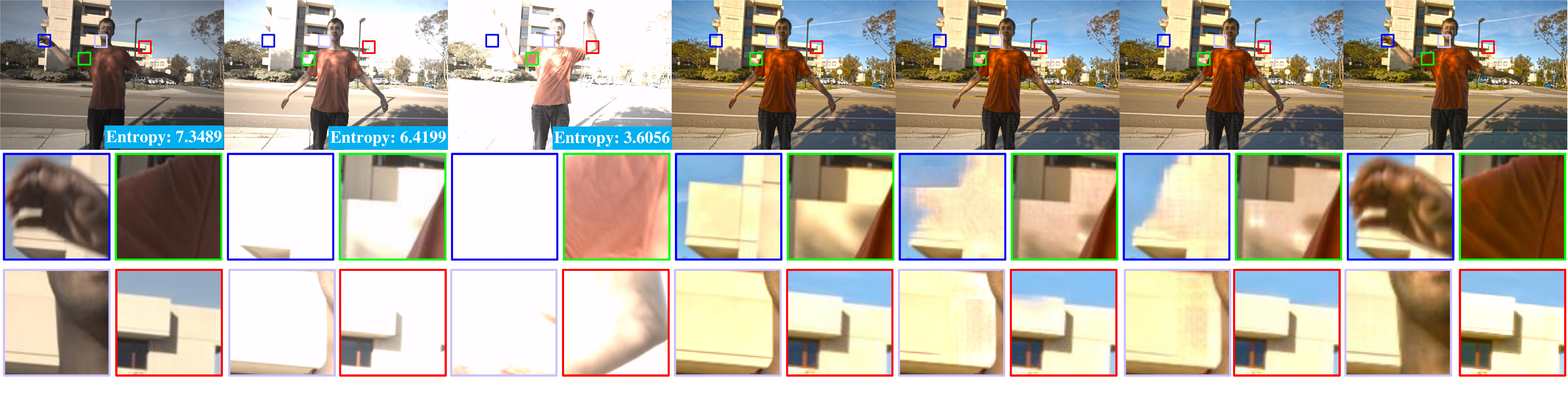}
		\put(4, -1){\tiny{(a) Short}}
		\put(17, -1){\tiny{(b) Middle}}
		\put(32, -1){\tiny{(c) Long}}
		\put(47, -1){\tiny{(d) GT}}
		\put(57, -1){\tiny{ (e) CA-ViT \cite{liu2022ghost}}}
		\put(72.5, -1){\tiny{(f) PASTA-I}}
		\put(83.5, -1){\tiny{(g) PASTA-I (ref. a)}}
	\end{overpic}
	\caption{Twp types of occlusion, \ie, compensatory and non-compensatory occlusion (NCO). The former denotes the provision of beneficial information in the corresponding regions between frames, \eg, the enlarged red region, while the later the simultaneous presence of occlusion and saturation in the corresponding regions between frames \eg, the first three close-up regions. NCO can be effectively utilized by re-selecting the reference frame from the perspective of maximizing information (\eg, entropy), rather than attempting to hallucinate or fill in the occluded regions. Among them, ``PASTA-I (ref. a)'' is generated using the proposed method, where ``short'' is selected as the \textit{reference frame} with entropy-maximization in the exposure sequence. This strategy ensures a high-fidelity result without ghosting.}
	\label{fig:occlusion_and_saturation}
	\vspace{-4mm}
\end{figure*}
\vspace{-4mm}
\subsection{Discussion}
\noindent{\bf Occlusion Coexist with Saturation.} In both our method and CA-ViT \cite{liu2022ghost}, pixel recovery encounters hurdles when occlusions coincide with saturation, as depicted in \cref{fig:occlusion_and_saturation}, mirroring challenges observed in human perception. Despite these obstacles, our model outperforms CA-ViT, generating more realistic estimates. In fact, the effective generation of high-fidelity HDR images without ghosting benefits from neighboring frames compensating for occlusions, illustrated in the pink and red regions. Complications typically arise when the corresponding region in neighboring frames is saturated, hindering realistic content hallucination, as seen in the blue and green areas. While techniques like diffusion models \cite{ho2020denoising,fei2023generative} or inpainting may be applied, they may not produce realistic results. For such case, leveraging occlusion rather than attempting to hallucinate could be a practical approach. Opting for a frame with maximum information content (\eg, entropy) as the reference from the exposure sequence can be advantageous. Success with image entropy in measuring pixel uncertainty \cite{jacobs2008automatic} supports this strategy. When constructing datasets or developing unsupervised models, prioritizing frames with maximum information content as references may alleviate occlusion effects.

\noindent{\bf Potential Drawback to Benchmarks.} 
Despite some SOTA methods achieving high PSNR-\emph{l} values on benchmarks \cite{kalantari2017deep,tel2023alignment,liu2023joint}, significant deviations to around 30dB are observed in certain test cases. This discrepancy is attributed to non-compensatory occlusions and dense subtle motion, challenging pixel-wise evaluation metrics. Wind-induced subtle motion during outdoor shooting, impacting scenes like tree branches, leaves, grass, clouds, and flowing water, can have a substantial effect. This requires a further in-depth reflection on the GT generation approach.  

\section{Conclusions}

This work proposes a flexible and efficient framework, PASTA, to address addresses HDR deghosting's computational challenges, which optimizes the trade-off between performance and computational efficiency, particularly at high resolutions. By integrating hierarchical feature decoupling and progressive aggregation, PASTA significantly boosts inference speed, achieving a significant threefold improvement, and can even reach a ninefold increase without sacrificing image quality. The comprehensive experiments validate that PASTA not only rivals but surpasses the existing SOTA methods in performance metrics, all while streamlining the use of computational resources.

\clearpage  

\bibliographystyle{splncs04}
\bibliography{main}
\end{document}